\documentclass{article}
\usepackage{spconf,amsmath,graphicx}
\usepackage{subfig}
\usepackage{amsfonts}
\usepackage{verbatim}
\title{Pixel Level Data Augmentation for Semantic Image Segmentation using Generative Adversarial Networks}
 
\name{Shuangting Liu$^{\dagger}$ \qquad Jiaqi Zhang$^{\dagger}$ \qquad Yuxin Chen$^{\dagger}$ \qquad Yifan Liu$^{\dagger}$ \qquad Zengchang Qin$^{\dagger\ddagger}$ \qquad Tao Wan$^{\star}$ \thanks{Corresponding to: \{zcqin,wantao\}buaa.edu.cn}}
\address{$^{\dagger}$  Intelligent Computing and Machine Learning Lab, School of ASEE, Beihang University, China\\
$^{\ddagger}$Keep Labs, Keep Inc
$^{\star}$ School of Biological Science and Medical Engineering, Beihang University\\
}

\begin{document}
%\ninept
%
\maketitle
\begin{abstract}
Semantic segmentation is one of the basic topics in computer vision, it aims to assign semantic labels to every pixel of an image. Unbalanced semantic label distribution could have a negative influence on segmentation accuracy. In this paper, we investigate using data augmentation approach to balance the semantic label distribution in order to improve segmentation performance. 
We propose using generative adversarial networks (GANs) to generate realistic images for improving the performance of semantic segmentation networks. Experimental results show that the proposed method can not only improve segmentation performance on those classes with low accuracy, but also obtain 1.3\% to 2.1\% increase in average segmentation accuracy. It shows that this augmentation method can boost accuracy and be easily applicable to any other segmentation models.

\end{abstract}
\section{Introduction}
\label{sec:intro}
Semantic segmentation aims to assign semantic labels to every pixel of a given image, it is one of the basic tasks in computer vision.
Recently, many deep learning models \cite{FCN,maskRCNN,D9,D10} are proposed and have achieved great performance on this task. However,  
deep learning based segmentation is always data-hungry and needs huge amount of fine pixel-level labeled data, which are always hard to collect, not to mention the fact that manual annotations may have a huge cost.
Most previous work focus on improving the structure of deep neural networks (e.g. adding more layers \cite{Chen04}) to improve the accuracy, yet this approach can only improve average segmentation accuracy. In some practical applications, we may need to improve the performance of some specific classes as they may contain critical information. 
Unbalanced label distribution is one of the reasons causing low performance in segmentation. 
%This kind of low accuracy probably caused by the unbalanced dataset. 
Using data augmentation for enlarging training set can yield better results and that has reported in various literature \cite{Krizhevsky,Ciresan05,Simard07}.

In this research, we consider using Generative Adversarial Networks (GANs) \cite{Goodfellow08} for data augmentation in order to improve the segmentation accuracy. GANs are well used in computer vision and image processing \cite{Isola09,Ledig10} for generating \emph{realistic} images
by learning true label distribution in a zero-sum game framework.
Realistic image generation is a kind of image-to-image translation. 
The goal is using a semantically labeled image to generate a photographic image. 
Several methods have been proposed for this task including cascaded refinement networks \cite{H5}, conditional GANs \cite{pix2pixHD}, and semi-parametric synthesis \cite{semi}. \\

Data augmentation is simply the extension of the training data with generated data. Existing data augmentation techniques can roughly fall into two
following categories: (a) geometric transformation which is computationally cheap and generic. (b) guided-augmentation or task-specific methods which using specific labels to generate image data \cite{E6}. In the case of image classification, some methods like
Affine \cite{Ciresan05}, elastic deformations \cite{Simard07}, patches extraction and RGB channels intensities alteration \cite{Krizhevsky} are all belong to this category.
However, these methods only lead to an image-level transformation which only change depth or scale of image. They can improve the robustness of neural network, but actually no help for dividing a clear boundary between data manifolds. These methods do not improve the label distribution which is determined by higher-level features. As for the second type of data augmentation, many complex manipulated augmentation schemes have been proposed in  fields such as scene text recognition \cite{G15}, text localization \cite{G10}, person detection \cite{G36}, and emotion classification \cite{Xinyue11}. They all demonstrate the great performance on synthetic data. 

Our work is most similar to \cite{Xinyue11}, in which GANs are used to improve classification accuracy on classes with imbalanced data. In this paper, in order to boost the performance in segmentation tasks,
we explore how to use GANs to generate supplementary data with pixel-level annotation labels and balance data-distribution within the dataset. 
The main contributions of this paper are as follows: (1) we propose a pipeline model for data augmentation by using GANs to generate supplementary data for  semantic segmentation. (2) We propose a new method for image augmentation at pixel-level. (3) We improve the data distribution and increase segmentation accuracy of both specific classes and on average. 

%===========================================================

\section{Methodology} 
\label{sec:method}
%In this section, the pipeline of our approach is described. We propose a new method for data augmentation on pixel level with briefly three stages of work.
\begin{figure}
\centering
\includegraphics[width=0.48\textwidth]{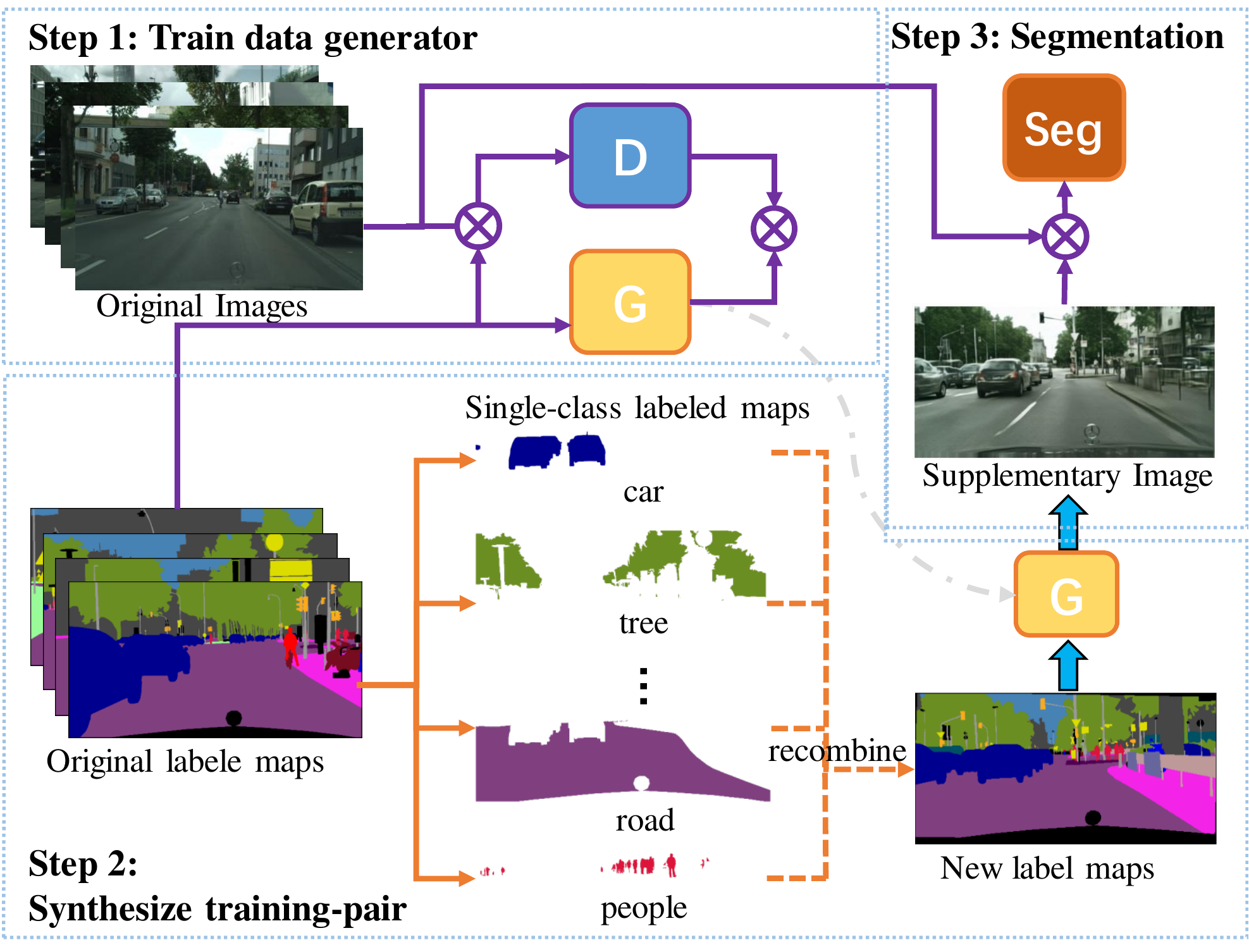}
\caption{Our pipeline model for data augmentation, where semantic labels \emph{car}, {tree} and \emph{people} can be used to reconstruct a new label map. }
\label{fig:method}
   
\end{figure}

%Our data augmentation pipeline is applicable to any pixel-level annotated datasets such as Cityscapes \cite{Cordts12}. Those datasets consist of real images and semantic label maps. Real images are captured by cameras in real scenes, as shown in Fig.~\ref{Fig:real}. Semantic label maps are corresponding images that have pixel-level annotation, as shown in Fig.~\ref{Fig:seg}. 

%Our data augmentation pipeline is applicable to any pixel-level annotated datasets such as Cityscapes \cite{Cordts12}. Those datasets consist of real images (Fig.~\ref{Fig:real}) and semantic label maps (Fig.~\ref{Fig:seg}).
%Real images are captured by cameras in real scenes, as shown in Fig.~\ref{Fig:real}. Semantic label maps are corresponding images that have pixel-level annotation, as shown in Fig.~\ref{Fig:seg}.
The main idea of our method is generating supplementary (augmented) data for semantic segmentation to balance the distribution of semantic labels and improve the segmentation results. Fig.~\ref{fig:method} schematically shows the procedure of the approach we proposed. The first step is training a GAN on original image/label pairs, it is used as a generator to transfer any human-designed semantic label maps (see Fig.\ref{fig:generation}b and \ref{fig:generation}c) to realistic images. In the second step, we use generated supplementary image to balance label distribution. Finally, we use supplementary data and original image data to train the segmentation network
for better segmentation results.

%------------------------------------------------------------------------- 
\subsection{Training Data Generator}
\label{sec:data-g}
\begin{figure}  
\centering
    \subfloat[]{
    \label{Fig:seg}
    \includegraphics[height=2cm]{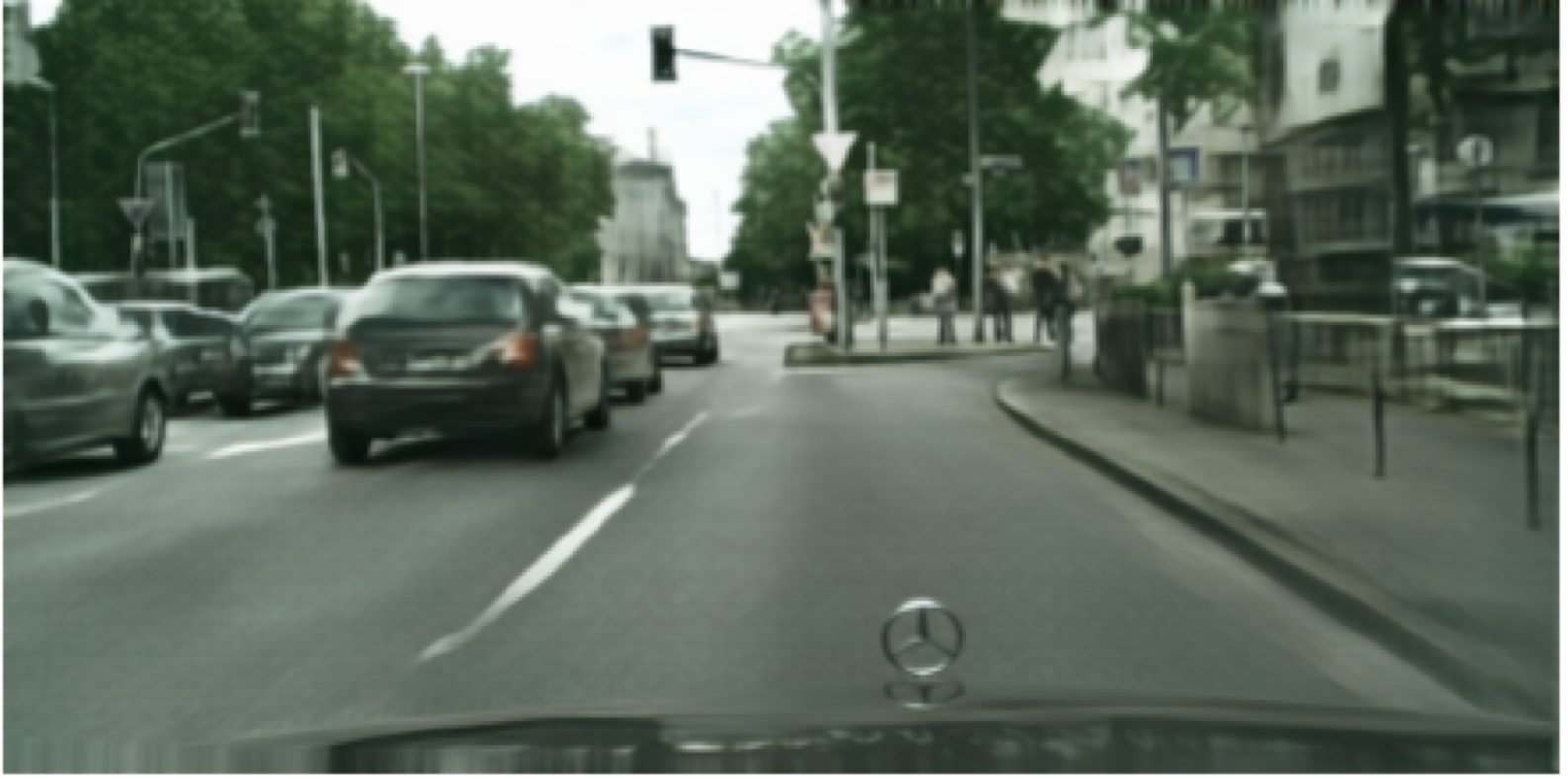}}
    \quad
    \subfloat[]{
    \label{Fig:real}
    \includegraphics[height=2cm]{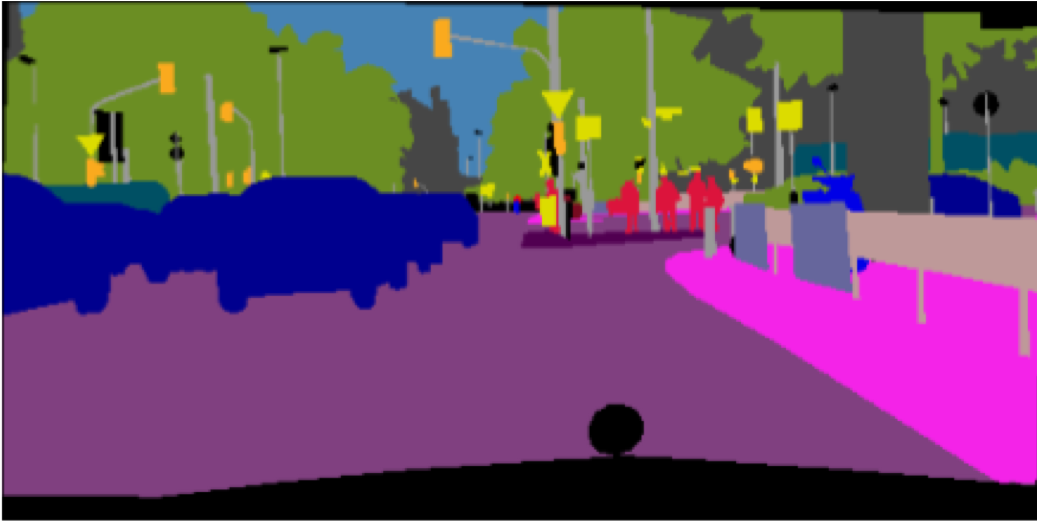}}
    \quad
    \subfloat[]{
    \label{Fig:seg1}
    \includegraphics[height=2cm]{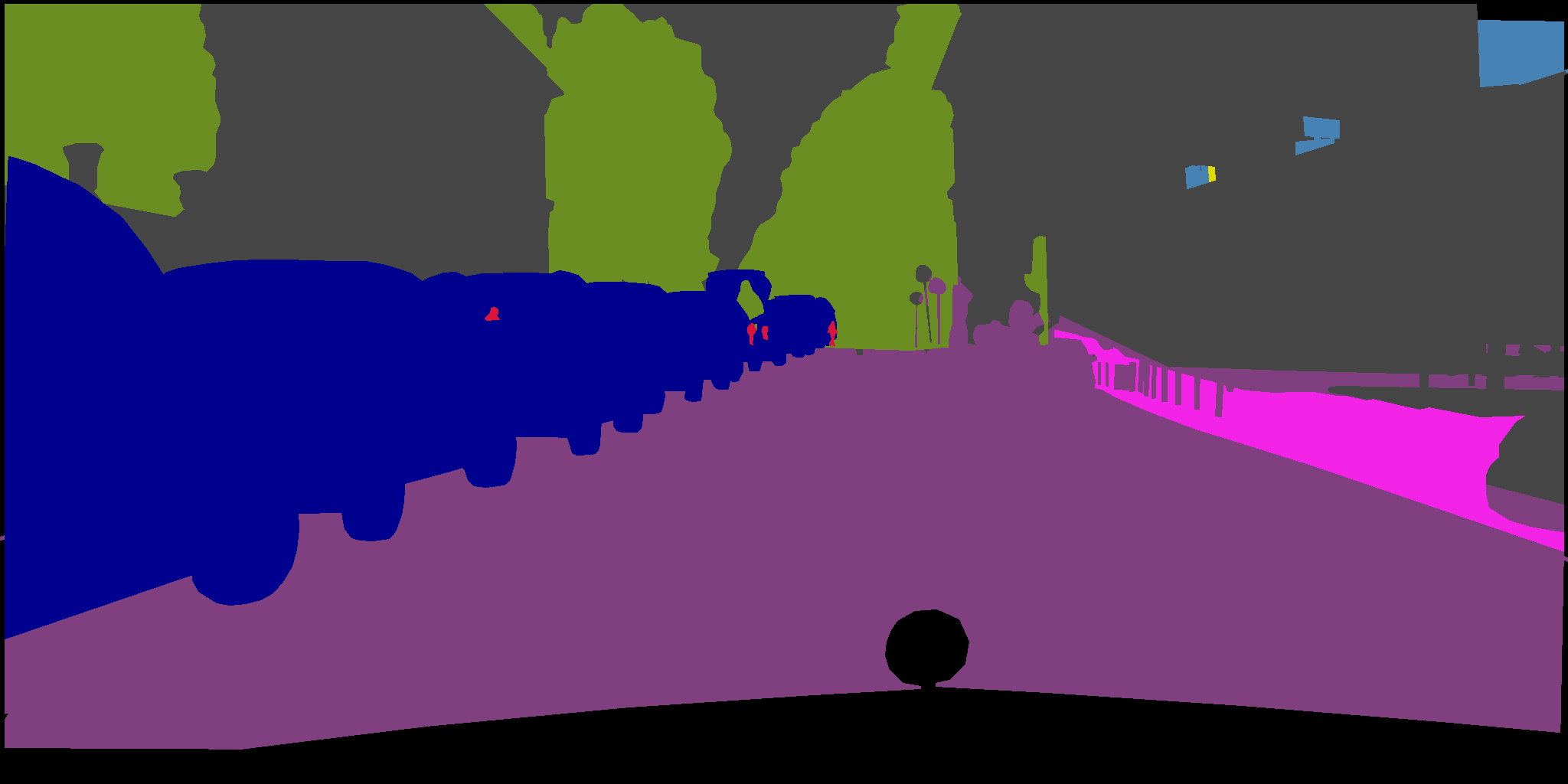}}
    \quad
    \subfloat[]{
    \label{Fig:real1}
    \includegraphics[height=2cm]{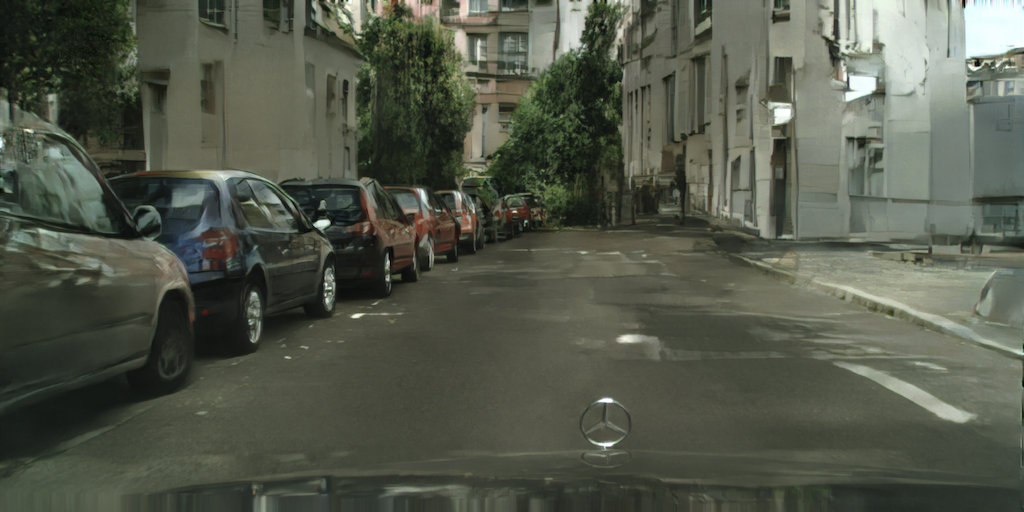}}\\
    \caption{An original image (a) and its corresponding semantic label map (b). We select several semantic labels including \emph{street}, \emph{car}, \emph{vegetation} and etc. to reconstruct a new semantic label map (c). Then we use GANs to generate its corresponding realistic image (d). }
    \label{fig:generation}

\end{figure}

We use the Pix2pix HD \cite{pix2pixHD} model to generate realistic images given a specific semantic label map as our data generator. Real images (e.g. Fig.~\ref{Fig:seg}) and their corresponding semantic label maps (e.g. Fig.~\ref{Fig:real}) from the original dataset are trained in pairs.
Besides the generator $G$, there is a discriminator $D$ to help completing the whole training process. $G$ and $D$ constitute Generative Adversarial Networks (GANs) \cite{Goodfellow08} . The aim of generator $G$ is to transfer semantic label maps to realistic images, while the discriminator $D$ is used to distinguish real images (original images) from fake and \emph{realistic} images generated by the generator $G$. We use minimax algorithm to model the strategy. 
\begin{equation}\min_{G}\max_{D}L_{GAN}(G,D)\label{equ:1}\end{equation}
%After the training process, we obtain a data generator which can transfer semantic label maps into realistic images.
%The training set is given a pair of images (${s}{_i}$,${r}{_i}$). ‘${s}{_i}$’ represents the semantic label map and ‘${r}{_i}$’ represents the corresponding real image. Using semantic label map ${s}{_i}$, G tries to generate image G(${s}{_i}$) that can fool discriminator to consider it as the real image ${r}{_i}$. Therefore,

%To further obtain high-resolution and realistic result, generator G is divided into 2 parts:{G1,G2}. Sub generator G1 is a global network which can extract global feature, while sub generator G2 is a local network which aims to synthesis more realistic details. To distinguish high-resolution real images with translated ones, three multi-scale discriminator is used in this framework. 

%The final function we aim to solve is:
%\begin{equation}\min_{G}((\max_{D_1,D_2,D_3}\sum_{k=1,2,3}{L_{GAN}(G,D_k))+\lambda\sum_{k=1,2,3}L_{FM}(G,D_k)})\label{equ:5}\end{equation}

%------------------------------------------------------------------------- 
\subsection{Synthesis of Training Data}
We create a new synthesis dataset based on reconstructed label maps and generate corresponding realistic images to train semantic segmentation networks.
%To synthesis new label maps which can balance the semantic label distribution, we can reconstruct objects (semantic labels) from the original dataset to form a new street scene.
To begin with, we separate each semantic label map in training set according to class of labels. Given a dataset of size $N$ with $K$ class of labels, we use ${I}{_1}$, ${I}{_2}$, ..., ${I}{_N}$ to represent corresponding semantic label maps of given original images. Separating those maps, we can then extract $m (m\leq K)$ semantic labels from each map. %, where $m$ is the number of labels contained in the image. 
We represent this process by:  ${I} \xrightarrow{} \left\{ {L}{_1}, {L}{_2}, ... , {L}{_m}\right\}$, ${L}{_i}\in\mathbf{L}$ and $|\mathbf{L}| = K$. 
Note that because one pixel only has one annotation, ${L}{_1}\cap{L}{_2}\cap ... \cap{L}{_m}=\phi$. In this way, one semantic label map can be separated into $m$ single label maps.
%And given a specific $c$ , ${P}{_i}{^c}$ $(i=1,2,...,N)$ are all the labels belong to class $c$.
%For those label classes that have instance label annotation in dataset, we further separate different object instances within the same category. For ${P}{_i}{^j}$, we can extract ${P}{_i}{^j}(1)$,${P}{_i}{^j}(2)$,...,${P}{_i}{^j}(q)$ as each object instance.

Then, we can reconstruct semantic label maps with these semantic labels. We arbitrarily select $n$ semantic labels $L_j \in \mathbf{L}$ for $j =1, \cdots, n$. Note that here, $L_j$ for $j =1, \cdots, n$ may come from different label maps, so they are not mutually exclusive. We combine them together to form a new label map (Fig.~\ref{Fig:seg1}). The process can be presented as following: $\left\{ {L}{_1}, {L}{_2}, ... , {L}{_n}\right\} \xrightarrow{} {R}$. We design several specific ways to reconstruct the new label maps (details are explained in experimental study). We can change the semantic label distribution by modifying the proportion of each label in the images. 
With the recombined label maps, we can generate realistic images (Fig.~\ref{Fig:real1}) by the data generator trained in Section \ref{sec:data-g}.
%Finally, we use trained dataset generator to generate realistic supplementary images (as is shown in Fig.~\ref{Fig:real1}). Using $Q_i$ (recombined label maps) as input. Dataset generator will generate a corresponding generated images $T_i (i=1,2,...,M)$, respectively. Each $Q_i (i=1,2,...,M)$ and corresponding $T_i (i=1,2,...,M)$ constitute training-pairs, (supplementary dataset), for further segmentation task. Therefore, we generate $M$ training-pairs as supplementary data based on $N$ original data.
%Such supplementary dataset can be used to balance the data distribution of original dataset because we can change the data distribution as we want in dataset. For example, if there are only a few wall-class label maps appearing in original dataset, we can add more 'Wall' in supplementary data.

\subsection{Segmentation Using Data Augmentation}

We use both original data and supplementary data (generated data) to train the semantic segmentation network. In the training process, we first train segmentation network with supplementary data. In this way, we can get a better initialization which contains the prior of some rare classes. 
Then, we use original dataset to fine tune the network. We use random initialization as our baseline. The comparison of two methods is discussed in experiment part.
The translation of supplementary images is shown in Fig.~\ref{fig:generation}.

\section{Experimental Studies}
\label{sec:typestyle}
In this section, we use PSPnet \cite{D81} as the segmentation model. We compare the model performance before and after the augmentation to verify the effectiveness of our method. 
%The PSPnet is first trained by original dataset, then followed by supplementary dataset. 
%\subsection{Datasets}

%------------------------------------------------------------------------- 
\subsection{Analysis of Semantic Label Distribution}

%Before doing experiments on Cityscapes dataset, we first analysis the distribution of data. 

%In the image-level labeled dataset such as emotion dataset or Mnist dataset, one image has one label. Thus, data distribution is mainly related to the number of times that images of each class appear in whole dataset. For example, in the FER2013 emotion dataset, the amount of ‘disgust’ and ’sad’ images are much less than others, which causes imbalanced data distribution. However, for pixel-level labeled dataset such as Cityscapes which is much more complex, each pixel has its own label. Thus, there are two main reasons that may influence data distribution and cause the imbalance of data: 1) The number of images in dataset that contain a specific class.(We call this as appearance frequency) 2) Amount of pixels of each class in one image. 

We choose Cityscapes \cite{Cordts12} as the test dataset. This dataset records city street scenes in 50 different cities. It defines 30 visual classes (labels) for annotation, leaving 19 classes for evaluation. In our experiments, we just use 19 classes of semantic labels.
\begin{figure}
\centering
\includegraphics[width=0.5\textwidth]{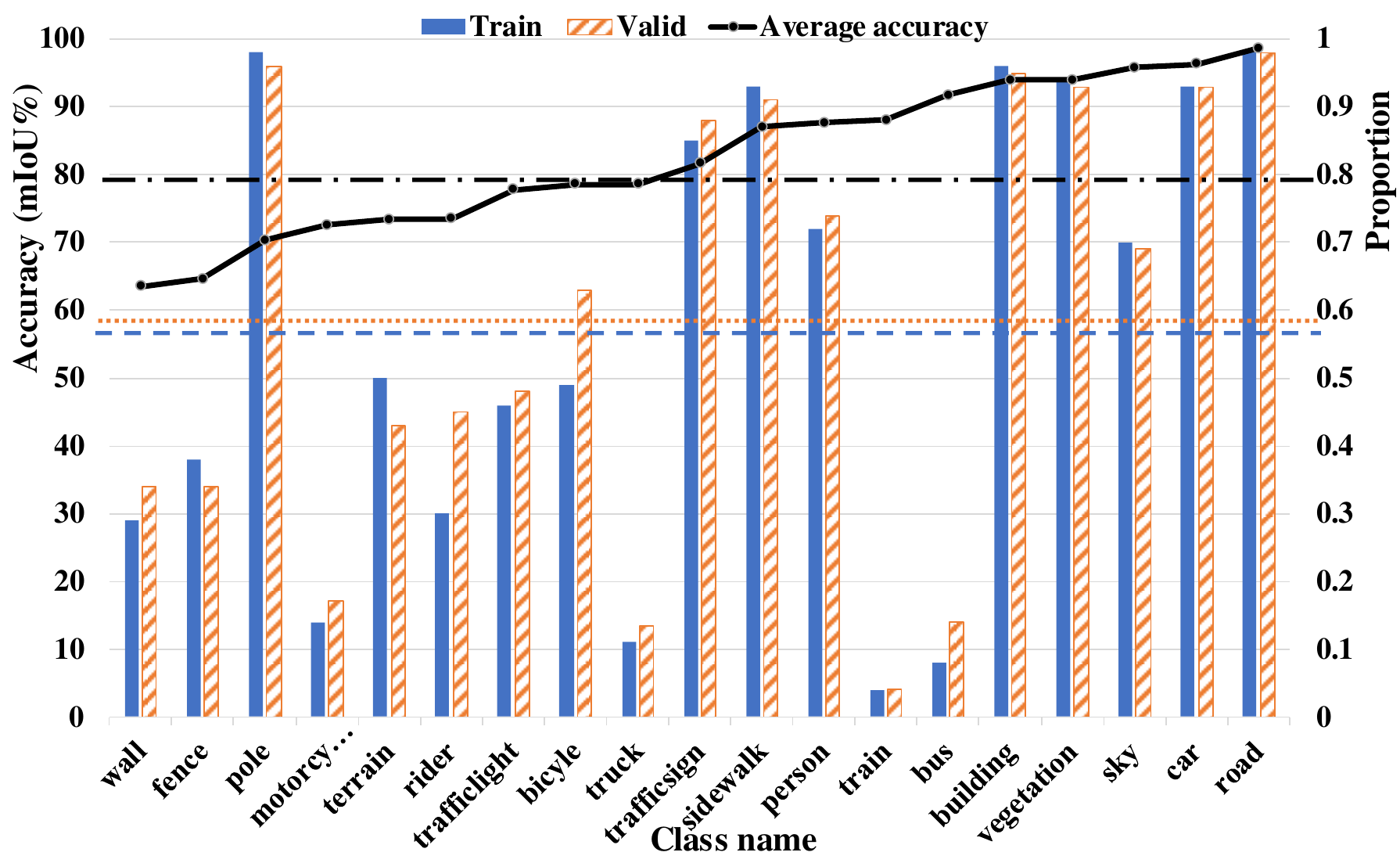}
\caption{Label distribution analysis and model accuracy.}
\label{fig:analysis}
\end{figure}
We first calculate the label distribution. For each label class, the frequency of each label class appearing in the training set and the validation set is derived, and we call it as \emph{appearance frequency}. Then we calculate the average segmentation accuracy of top 5 ranked models on Cityscapes website. Fig.~\ref{fig:analysis} illustrates the correlation between the label distribution and segmentation accuracy.
Comparing those classes with low appearance frequency and those have low segmentation accuracy, we find out that two groups are highly overlapped. In other word, it is possible to balance data distribution and further improve segmentation accuracy by increasing the appearance frequency on some specific classes.

\begin{figure}  
\centering
    \subfloat[]{
    \label{fig:labelMAP}
    \includegraphics[height=2cm]{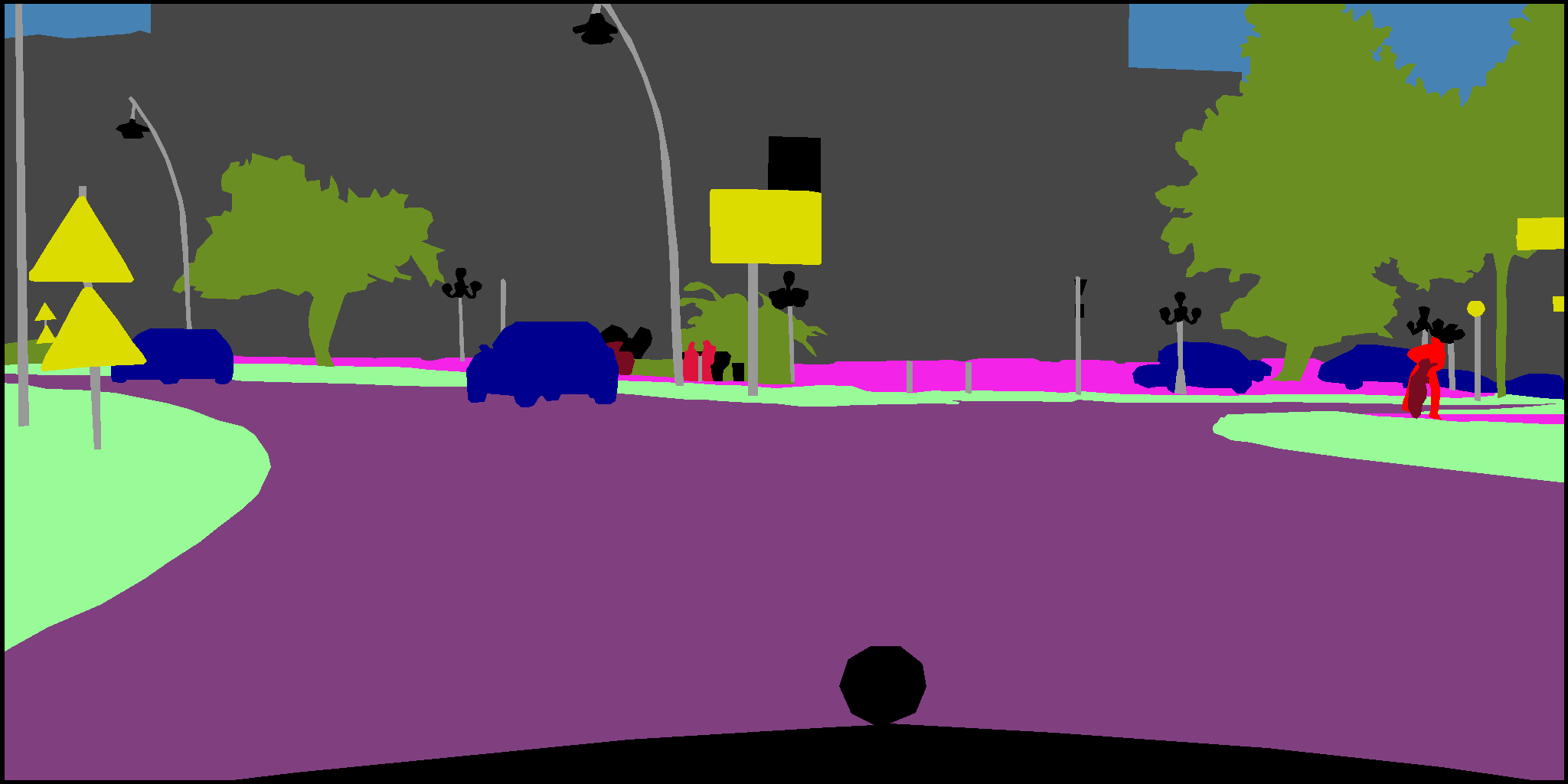}}
    \quad
    \subfloat[]{
    \label{fig:addwall}
    \includegraphics[height=2cm]{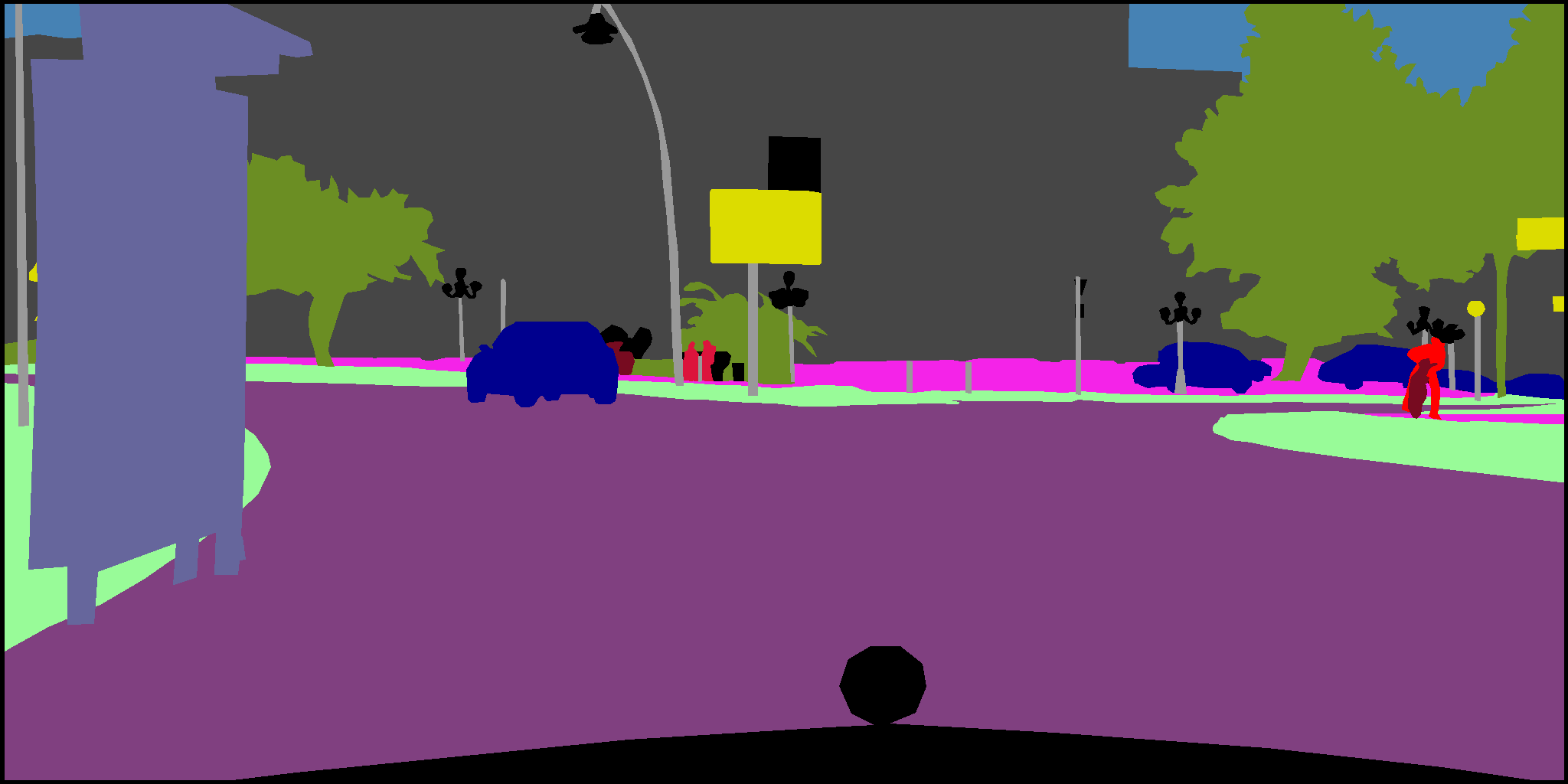}}
    \quad
   \\

    \caption{
    \protect\subref{fig:labelMAP} Original label map and \protect\subref{fig:addwall} the new label map by adding a semantic label \emph{wall} (grey area on the left).}
    \label{fig:single}
\end{figure}

\subsection{Ablation Study}

According to the analysis of label distribution, we propose two ways to obtain new labels: 1) Overlay a single label  directly on original label maps from the dataset. 2) Totally reconstruct using labels to form an entirely new label map. 
% We design three different kinds of experiments to verify the effectiveness of our methods. 1) Add single class. 2) Add multiple classes. 3) Reconstruct the whole image.
To further study how the proportion of supplementary data in training data will influence the performance of semantic segmentation network, we conduct experiments with different proportion of augmentation data.

\begin{figure}
   \centering
   \includegraphics[width=0.5\textwidth]{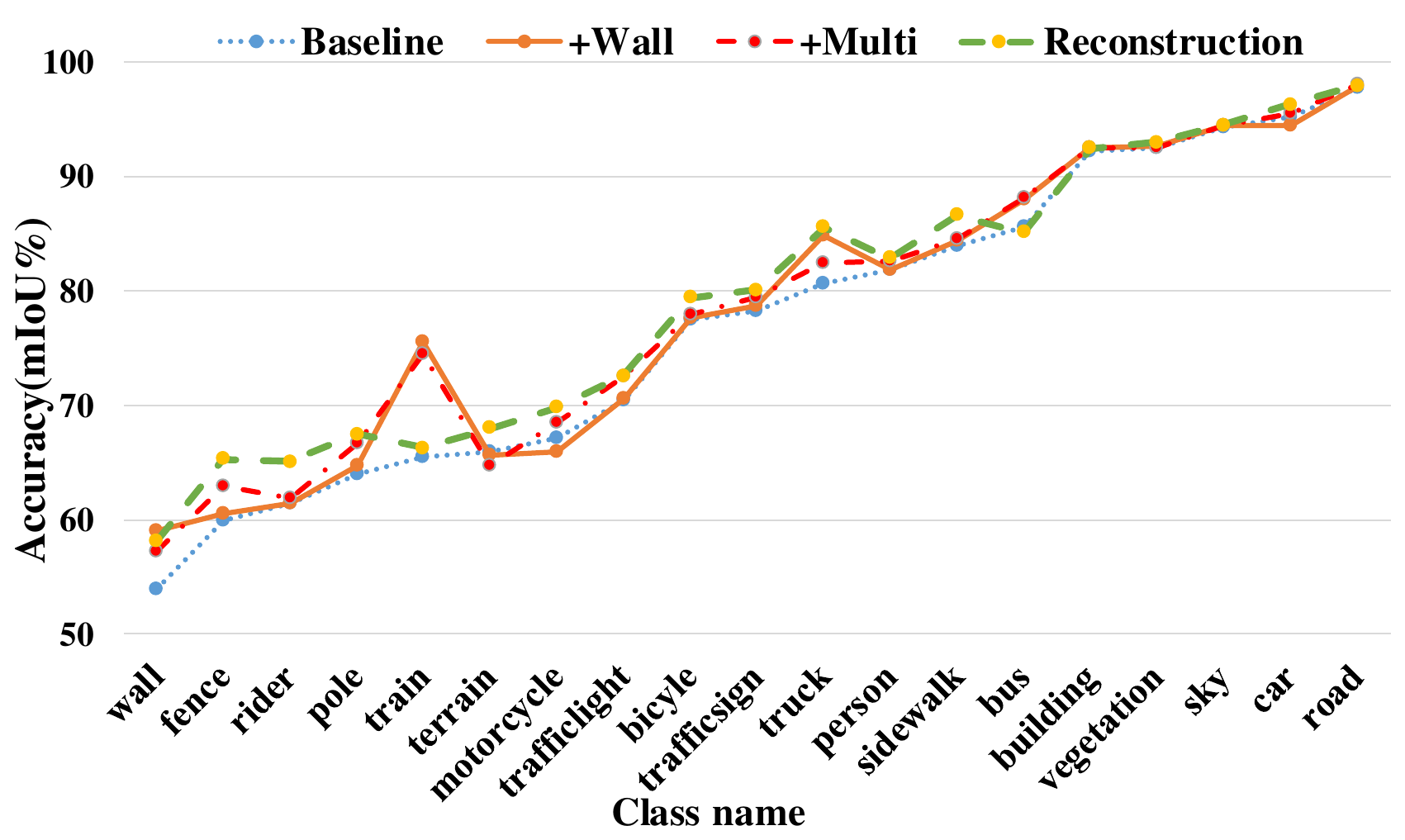}
   \caption{Results comparison. Baseline: only use original Cityscpaes dataset. +Wall: add label \emph{wall} to each image. +Multi: +Wall +Fence +Pole +Traffic light +Train. Reconstruction: a new label map. 
   %(Supplementary data takes up 50\% in all training dataset.)
   }
   \label{fig:result}
\end{figure}

\noindent\textbf{Overlaying single label} We start with adding one label on original label maps to verify the effectiveness of our method.
We increase the appearance frequency of those classes with low segmentation accuracy by overlaying specific single label on original label maps that do not contain this class. Fig.~\ref{fig:single} shows the label map before and after applying our method. Taking the class \emph{Wall} as an example, we first pick up all original label maps without the label class \emph{Wall} from training set. We randomly choose from all wall-class label maps, and overlay one wall-class label on each original map. Furthermore, we use GANs to transfer segment images into a translated image. Finally, we use these supplementary images and original images together in different proportion to train the semantic segmentation network and calculate intersection over union (IoU) of all the classes. 

We then study the effectiveness of adding more semantic labels on original label maps. We select several classes with low segmentation accuracy, and randomly choose some labels to add on. Single label of each class is overlaid on original label map. Supplementary images are also generated by GANs. In this experiment, we design one combination way that add \emph{Wall}, \emph{Fence}, \emph{Pole}, \emph{Traffic Light}, and \emph{Train}.
%For those classes like pole that nearly appear in every image but still have a low classification accuracy, we add the class 5 times in each image. It is not hard to understand. For most classes, having higher appearance frequency always means that having larger amounts of pixels in dataset. However, poles are too slender in one image, which means that poles take up small proportion pixels among all pixels in the dataset. In this case, only increasing the appearance frequency is not enough. Therefore, we add more poles in one image to increase the proportion of poles’ pixels. After augmenting dataset with images that add a single class, mean IoU and the IOU of added class increase to some extent. Specifically, the mean IoU increases 1.2\%-1.3\% and the IOU of added class increases 5.0\%-7.1\%. This result shows the effectiveness of our method in data augmentation. 
Results are shown in Fig.~\ref{fig:result} and Table ~\ref{table:ourmethod}. Adding one class of label improves mean IoU 1.3\% and the IoU of the added class increases 5.0\%. After adding multiple classes of labels, IoU increase significantly. Meanwhile, the mean IoU increases further up to about 1.5\%.
\begin{figure}  
\centering
    \subfloat[]{
    \label{fig:basic}
    \includegraphics[height=2cm]{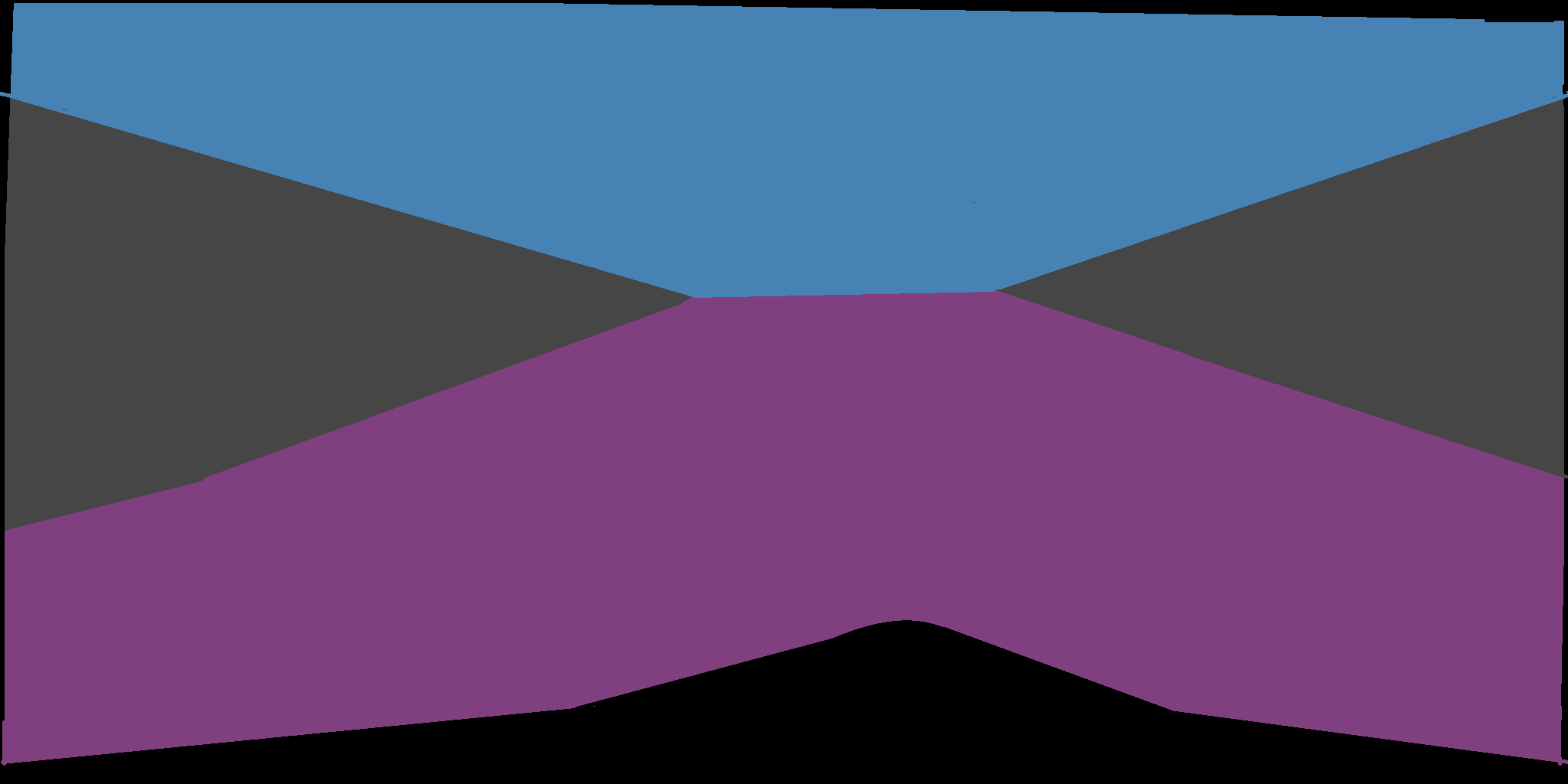}}
    \quad
    \subfloat[]{
    \label{fig:recombine}
    \includegraphics[height=2cm]{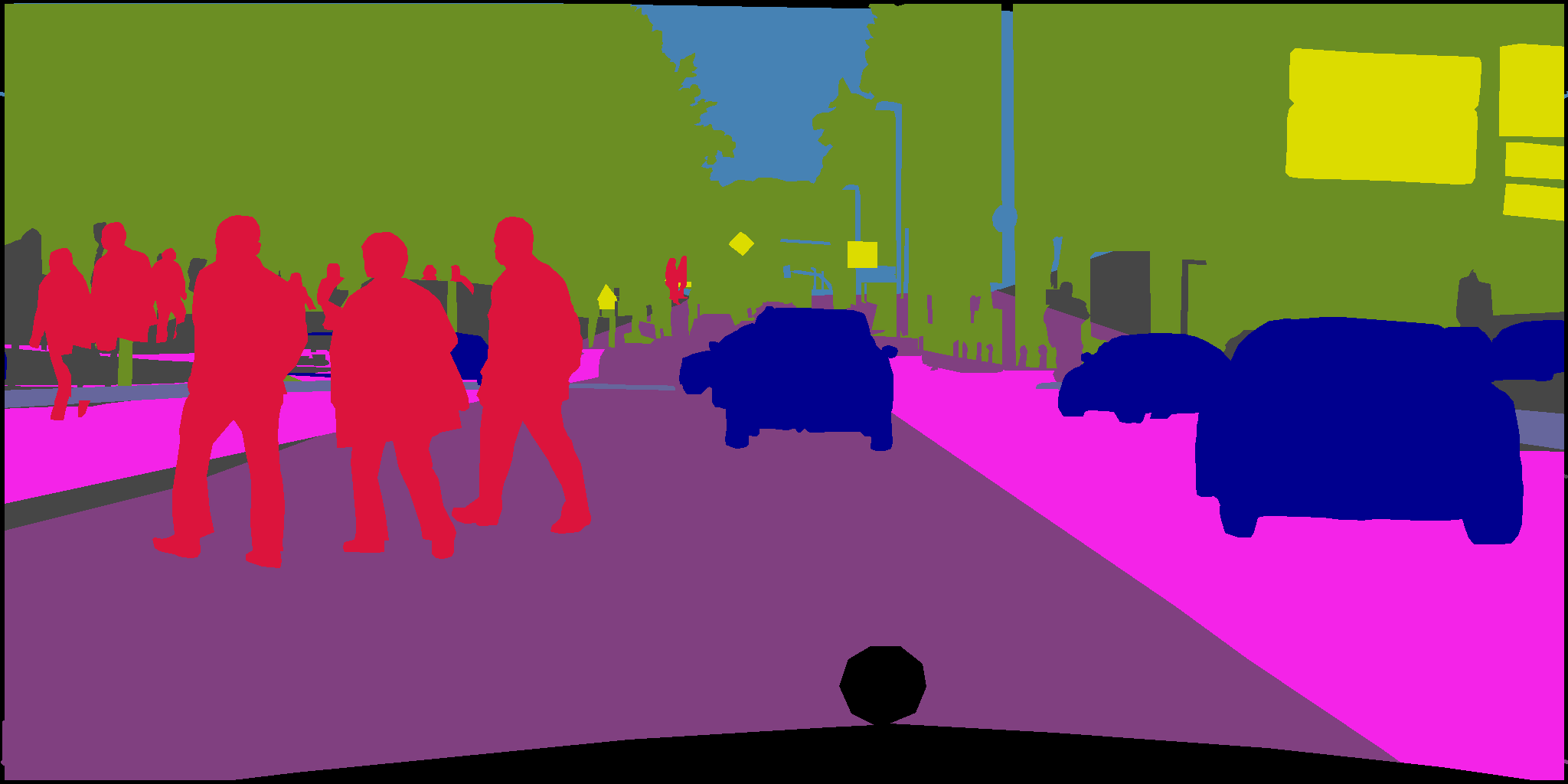}}
    \quad
   \\

    \caption{Reconstruction of label maps:
    \protect\subref{fig:basic} A basic label map and \protect\subref{fig:recombine} reconstructed label map by adding various semantic labels.}
    \label{fig:reconstruct}
\end{figure}

\noindent\textbf{Reconstruction} In order to balance dataset better, we pick up segmented images from each class and combine them together to form a totally new image. We first draw a basic label map which only contains the class label \emph{Sky}, \emph{Road} and \emph{Building}, which is shown in Fig.~\ref{fig:basic}. In this way we can make sure every pixel on the new label map has its label. Otherwise we may obtain an image with blank space labeled 0. Then we overlay each single label map on basic labels to form a constructed label map, as shown in Fig.~\ref{fig:recombine}. We use GANs to obtain corresponding translated images. We repeat the preceding procedure for 2 times. Two different datasets are generated and input in segmentation network, and the mean IoU increases 2.0\% and 2.1\%, respectively. Since two results are very similar, we just show one of results in Fig.~\ref{fig:result} and Table \ref{table:ourmethod}.
\setlength{\tabcolsep}{6.0pt}
\begin{table}
\setlength{\abovecaptionskip}{0.cm}
	\begin{center}
		\caption{
			Experimental results of our approaches.
		}
		\label{table:ourmethod}
		\begin{tabular}{cccc}
			\hline\noalign{\smallskip}
			Method & Single Label & Multi-Label & Reconstruction \\
			\noalign{\smallskip}
			\hline
			\noalign{\smallskip}
			mIoU & 78.65 &78.82 & 79.41 \\
		
			\noalign{\smallskip}
			\hline
		\end{tabular}
	\end{center}
\end{table}

\begin{figure}
	\centering
	\includegraphics[height=4.7cm]{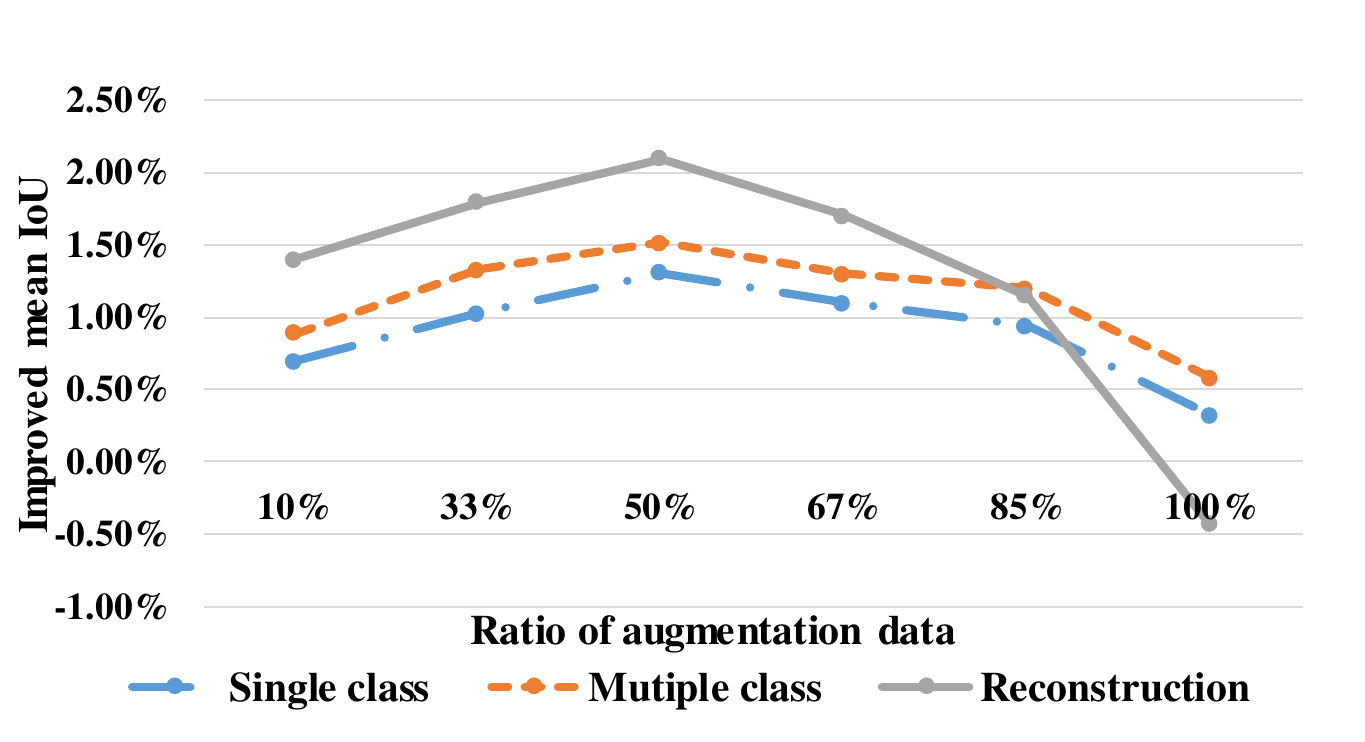}
	\caption{Influence of supplementary ratio in training set.}
	\label{fig:Ratio}
\end{figure}

\noindent\textbf{Ratio of supplementary data}
To figure out how the ratio of supplementary data and original data will influence segmentation accuracy, we did further experiment as shown in Fig~\ref{fig:Ratio}. Notice that when the proportion of supplementary data increasing, segmentation accuracy improves at first. But when the proportion continue goes up, the accuracy goes down. Results show that the best proportion of supplementary data in the training dataset is between 30\% to 70\%.

%-------------------------------------------------------------------------
\subsection{Model Comparisons and Analysis}
\begin{figure}  
\centering
    \subfloat[]{
    \label{Fig:transfer_orgi}
    \includegraphics[height=2cm]{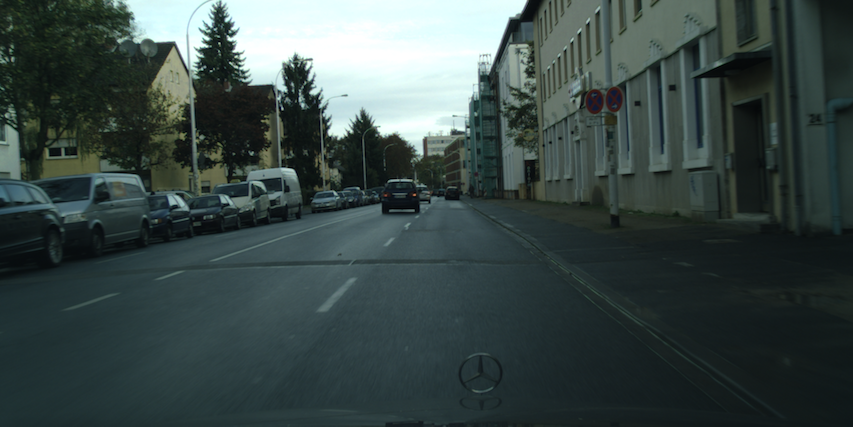}}
    \quad
    \subfloat[]{
    \label{Fig:transfer}
    \includegraphics[height=2cm]{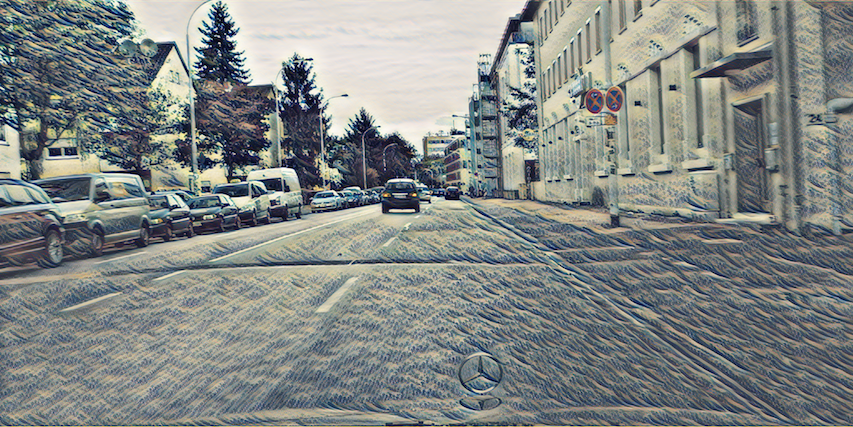}}
    \quad
   \\

    \caption{The result of generated an image using style transfer.
    \protect\subref{Fig:seg} the original image \protect\subref{Fig:real} the transferred image.}
    \label{fig:trans}
\end{figure}

We compare each of our data augmentation methods to traditional data augmentation methods using rotation and zooming in and a state-of-art data augmentation method using style transfer. The result of using style transfer is shown in Fig.~\ref{fig:trans}. Results of each method are shown in Table \ref{table:comparison}. According to the table, all of our methods have a better performance than traditional augmentation method, and our method outperforms state-of-art methods when using reconstructed data as supplementary data. Also, reconstructing separate classes can obtain highest segmentation accuracy among all our methods. This is because that it generates supplementary data with more variety and better balance the dataset. The results show the effectiveness of our method.
 
\setlength{\tabcolsep}{4.0pt}
\begin{table}[!h]
\setlength{\abovecaptionskip}{0.cm}
	\begin{center}
		\caption{
			Comparisons to other data augmentation models.
		}
		\label{table:comparison}
		\begin{tabular}{cccc}
			\hline\noalign{\smallskip}
			Method & Tradition & Style Transfer & Reconstruction \\
			\noalign{\smallskip}
			\hline
			\noalign{\smallskip}
			mIoU/Improve & 77.31 & 79.10/+1.79 & 79.41/+2.1 \\
			\noalign{\smallskip}
			\hline
		\end{tabular}
	\end{center}
\end{table}
\vspace{-2.5em}
%------------------------------------------------------------------------- 

\section{Conclusions}
\label{sec:conclusions}
In this paper, we explored how data augmentation method can be used to improve the performance of semantic segmentation. We proposed an augmentation method to generate supplementary data by using GANs. 
By adding generated label maps and images to original images 
%or recombining label classes together 
as supplementary data, we can improve the diversity of data and balance the semantic label distribution. Comparing to other approaches in experiments, we found that the best way to implement our approach was using the \emph{Reconstruction} method and setting the proportion of supplementary data as around 50\%. The results shown that mean accuracy of a specific class can increase up to 5.5\% and the average segmentation accuracy can increase 2\%.

%------------------------------------------------------------------------- 

\vfill\pagebreak

% References should be produced using the bibtex program from suitable
% BiBTeX files (here: strings, refs, manuals). The IEEEbib.bst bibliography
% style file from IEEE produces unsorted bibliography list.
% -------------------------------------------------------------------------
\bibliographystyle{IEEEbib}
\bibliography{strings,refs}

\end{document}